\documentclass[runningheads]{llncs}
\usepackage{graphicx}
\usepackage{amsmath,amssymb} 
\usepackage{color}
\usepackage{subfig}	
\usepackage{booktabs}
\usepackage[skip=3pt]{caption}
\usepackage{multirow}
\usepackage{xspace}
\usepackage{setspace}
\usepackage{xcolor}

\begin{document}

\title{Domain Transfer for 3D Pose Estimation from Color Images without Manual Annotations} 
\titlerunning{Domain Transfer for 3D Pose Estimation} 

\makeatletter
\DeclareRobustCommand\onedot{\futurelet\@let@token\@onedot}
\def\@onedot{\ifx\@let@token.\else.\null\fi\xspace}

\def\eg{\emph{e.g}\onedot} \def\Eg{\emph{E.g}\onedot}
\def\ie{\emph{i.e}\onedot} \def\Ie{\emph{I.e}\onedot}
\def\cf{\emph{c.f}\onedot} \def\Cf{\emph{C.f}\onedot}
\def\etc{\emph{etc}\onedot} \def\vs{\emph{vs}\onedot}
\def\wrt{w.r.t\onedot} \def\dof{d.o.f\onedot}
\def\etal{\emph{et al}\onedot}
\makeatother


\author{Mahdi Rad\inst{1}\orcidID{0000-0002-4011-4729} \and
Markus Oberweger\inst{1}\orcidID{0000-0003-4247-2818} \and
Vincent Lepetit\inst{2, 1}\orcidID{0000-0001-9985-4433}}
%

\authorrunning{Rad \etal} 


\institute{Institute for Computer Graphics and Vision, Graz University of Technology, Graz, Austria \and
Laboratoire Bordelais de Recherche en Informatique, Universit\'{e} de Bordeaux, Bordeaux, France\\
\email{\{rad,oberweger,lepetit\}@icg.tugraz.at}}

\maketitle

\begin{abstract}
We introduce a novel learning method for 3D pose estimation from color images. While acquiring annotations for color images is a difficult task, our approach circumvents this problem by learning a mapping from paired color and depth images captured with an RGB-D camera. We jointly learn the pose from synthetic depth images that are easy to generate, and learn to align these synthetic depth images with the real depth images.
We show our approach for the task of 3D hand pose estimation and 3D object pose estimation, both from color images only. Our method achieves performances comparable to state-of-the-art methods on popular benchmark datasets, without requiring any annotations for the color images.

\keywords{Domain transfer \and 3D object pose estimation  \and 3D hand pose estimation \and Synthetic data.}
\end{abstract}

\section{Introduction}

\begin{figure}[t]
  \begin{center}
  \includegraphics[trim={3.3cm 3.5cm 2cm 3cm},clip,width=0.65\linewidth]{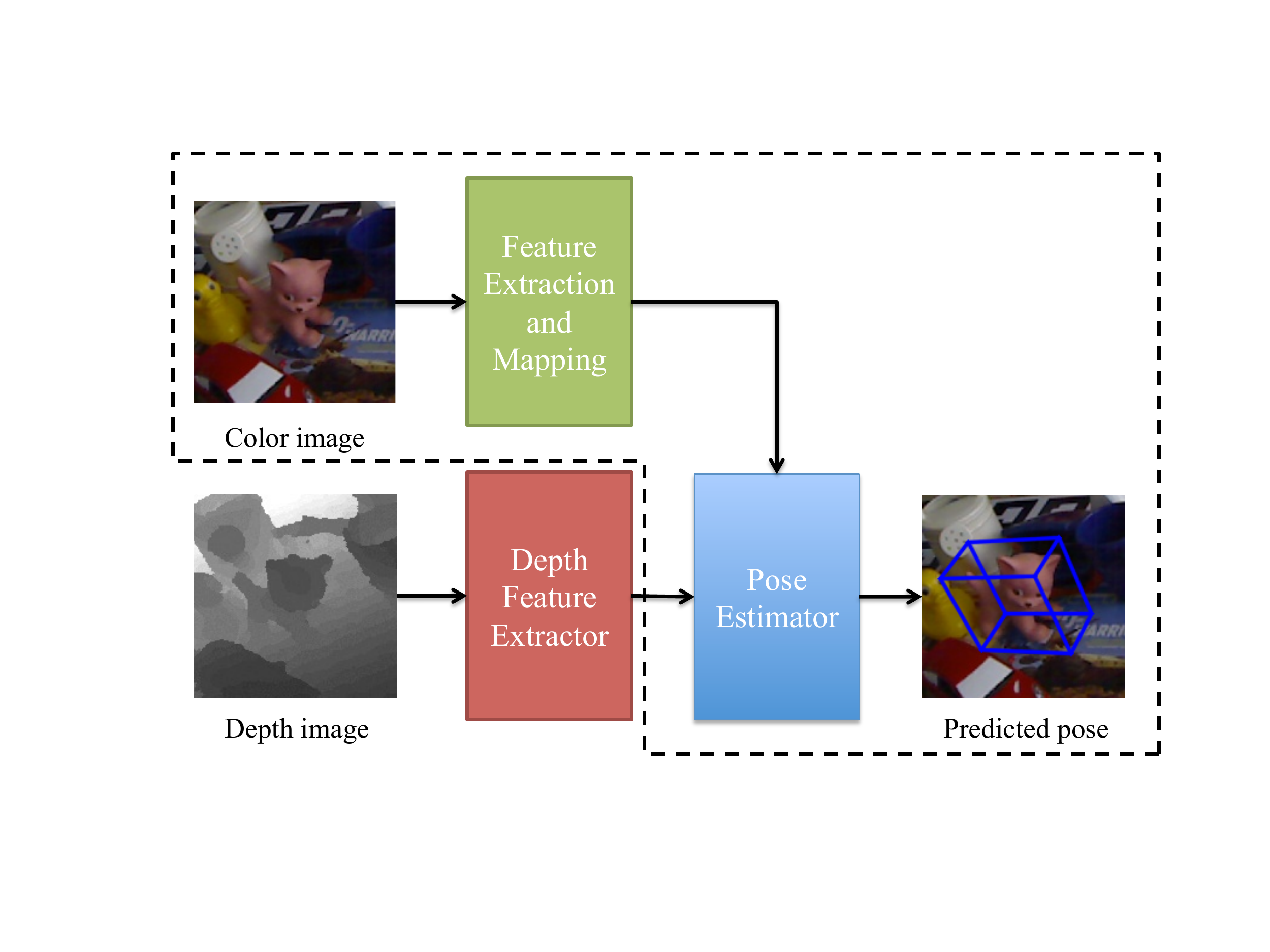} 
\end{center}

\caption{ \label{fig:overview} Method overview. We train a depth feature extractor (red box) together with a pose estimator (blue box). We also train a second network (green box), which extracts image color features and maps them to the depth space, given color images and their corresponding depth images.  At run-time, given a color image, we map color features to depth space in order to use the pose estimator to predict the 3D pose of the object (dashed lines).  This removes the need for labeled color images.}
\end{figure}

\begin{figure}[t]
  \begin{center}
  \begin{tabular}{cccc}
  \includegraphics[width=0.22\linewidth]{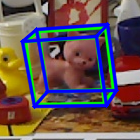}  &
  \includegraphics[width=0.22\linewidth]{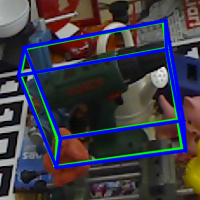}  &
  \includegraphics[width=0.22\linewidth]{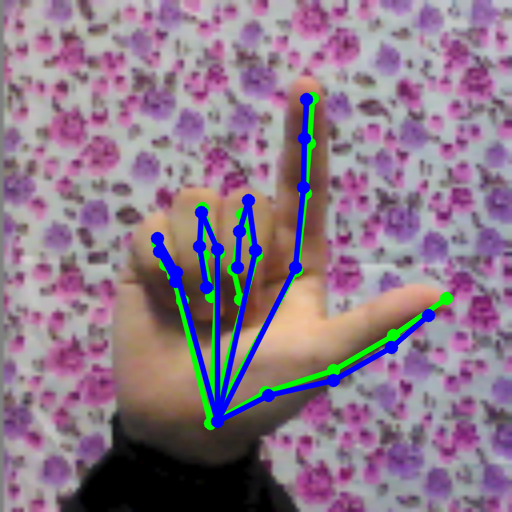}  &
  \includegraphics[width=0.22\linewidth]{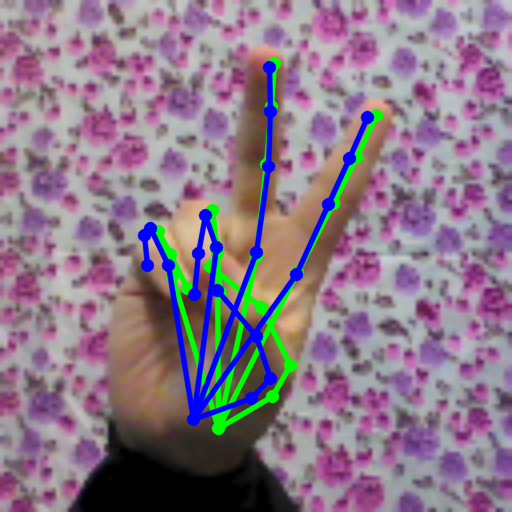}  \\

  \end{tabular}
\end{center}

\caption{ \label{fig:teaser}  Our method allows very accurate 3D pose estimation from color images without annotated color images. 
In case of 3D rigid object pose estimation we draw the bounding boxes, where blue is the ground truth bounding box and red the bounding box of our predicted pose. For 3D hand pose estimation, we show the 3D joint locations projected to the color image, blue denoting the ground truth and green our estimation.}
\end{figure}

3D pose estimation is an important problem with many potential applications.  Recently, Deep Learning methods have demonstrated great performance, when a large amount of training data is available~\cite{Xiang2018,Rad2018,Tekin2018,Rad17,Oberweger17}.  To create training data, the labeling is usually done with the help of markers~\cite{Hinterstoisser12b,Hodan17} or a robotic system~\cite{Calli2017}, which in both cases is very cumbersome, expensive, or sometimes even impossible, especially from color images. For example, markers cannot be used for 3D hand labeling of color images, as they change the appearance of the hand.

Another direction is to use synthetic images for training.  However, synthetic images do not exactly look like real images. Generative Adversarial Networks~(GANs)~\cite{Bousmalis16,Goodfellow14,Shrivastava16,Zhu17} or transfer learning techniques~\cite{Ganin15,Muandet13,Pan09,Rozantsev17} can be used to bridge the domain gap between real and synthetic images. However, these approaches still require some annotated real images to learn the domain transfer. \cite{Rad2018} relies on registered real images to compute a direct mapping between the image features of real and synthetic images, but it also requires some labeled real images.

In this paper, we propose a method that learns to predict a 3D pose from color images, without requiring labeled color images. Instead, it exploits labeled depth images. These depth images can be real depth images, which are easier to label than color images, and are already readily available for some problems.  More interestingly, they can also be synthetic depth images: Compared to color images, synthetic depth images are easier to render, as there is no texture or illumination present in these images.

An overview of our approach is shown in Fig.~\ref{fig:overview}.
Our main idea is to bridge the domain gap between color images and these synthetic depth images in two steps, each one solving an easier problem than the original one.  We use an RGB-D camera to capture a set of pairs made of color and depth images that correspond to the same view. Capturing such a set can be done by simply moving the camera around. We apply \cite{Rad2018} to this set and learn to map the features from the color images to corresponding depth images. However, this mapping alone is not sufficient: A domain gap between the depth images captured by the RGB-D camera and the available labeled depth images remains, since the labeled depth images could be captured with another RGB-D camera or rendered synthetically. Fortunately, this remaining gap is easier to bridge than the domain gap between real and synthetic color images, since illumination and texture effects are not present in depth images.  To handle it, we use Maximum Mean Discrepancy~(MMD)~\cite{Gretton08}
to measure and minimize the distance between the means of the features of the real and synthetic depth images mapped into a Reproducing Kernel Hilbert Space~(RKHS).
MMD is a popular in domain transfer method~\cite{Rozantsev17} since it does not require correspondences to align the features of different domains and can be efficiently implemented.

Our approach is general, and not limited to rigid objects. It can be applied to many other applications, such as 3D hand pose estimation, human pose estimation, etc.  Furthermore, in contrast to color rendering, no prior information about object's texture has to be known.  Fig.~\ref{fig:teaser} shows applications to two different problems: 3D rigid object pose estimation and 3D hand pose estimation from color images, on the LINEMOD~\cite{Hinterstoisser12b} and STB~\cite{Zhang2016} datasets, respectively.  Our method achieves performance comparable to state-of-the-art methods on these datasets without requiring any annotations for the color images.

In the remainder of this paper, we discuss related work, then present our approach and its evaluation.


\section{Related Work}

We first review relevant works for 3D pose estimation from color images, and then review related methods on domain transfer learning.

\subsection{3D Pose Estimation from Color Images}
Inferring the 3D pose from depth images has achieved excellent results~\cite{Xiang2018,Oberweger17,Mueller17,Krull15}, however, inferring the 3D pose from color images still remains challenging.  \cite{Rad17} presented an approach for 3D object pose estimation from color images by predicting the 2D locations of the object corners and using P$n$P to infer the 3D pose, similar to~\cite{Tekin2018,Oberweger2018}.  Also, \cite{Zimmermann2017} first predicts the 2D joint locations for hand pose estimation, and then lifts these prediction to 3D estimates. \cite{Mueller2018} predicts 2D and 3D joint locations jointly, and then applies inverse kinematics to align these predictions. Similarly, \cite{Panteleris2018} uses inverse kinematics to lift predicted 2D joint locations to 3D. All these approaches are fully supervised and require annotated color images, which are cumbersome to acquire in practice.  Recently, \cite{Kehl17} uses synthetically generated color images from 3D object models with pretrained features, however, they require extensive refinement of the initial network predictions, and we will show that we can reach better performances without annotations for real color images when using no refinement. To generalize synthetically generated color images to real images, \cite{sundermeyer2018implicit} proposed to use a domain randomization method, however, the generalization is still limited, and outperformed by our approach as we show in the Evaluation section.

\subsection{Domain Transfer Learning}
As we mentioned in the introduction, it is difficult to acquire annotations for real training data, and training on synthetic data leads to poor results~\cite{Rad17,Kehl17}. This is an indication for a domain gap between synthetic and real training data. Moreover, using synthetic data still requires accurately textured models~\cite{Rad17,Calli2017,Kehl17,Hinterstoisser17} that require large amount of engineering to model. On the other hand, synthetic depth data is much simpler to produce, but still it requires a method for domain transfer.

A popular method is to align the distributions for the extracted features from the different domains. Generative Adversarial Networks~(GANs)~\cite{Goodfellow14} and Variational Autoencoders~(VAEs)~\cite{Kingma2014} can be used to learn a common embedding for the different domains. This usually involves learning a transformation of the data such that the distributions match in a common subspace~\cite{Ganin15,Muandet13,Pan09}. \cite{Spurr2018} learns a shared embedding of images and 3D poses, but it requires annotations for the images to learn this mapping.
Although GANs are able to generate visually similar images between different domains~\cite{Zhu17}, the synthesized images lack precision required to train 3D pose estimation methods~\cite{Rad2018,Bousmalis16}. Therefore, \cite{Mueller2018} developed a sophisticated GAN approach to adapt the visual appearance of synthetically rendered images to real images, but this still requires renderings of high-quality synthetic color images.

To bridge this domain gap, \cite{Rad2018} predicts synthetic features from real features and use these predicted features for inference, but this works only for a single modality, \ie depth or color images, and requires annotations from both modalities.  Similarly, \cite{Gupta2016} transfers supervision between images from different modalities by learning representations from a large labeled modality as a supervisory signal for training representations for a new unlabeled paired modality. In our case, however, we have an additional domain gap between real and synthetic depth data, which is not considered in their work.  Also, \cite{Cai2018} aims at transforming the source features into the space of target features by optimizing an adversarial loss.  However, they have only demonstrated this for classification, and this approach works poorly for regression~\cite{Rad2018}.  \cite{Rozantsev17} proposed a Siamese Network for domain adaptation, but instead of sharing the weights between the two streams, their method allows the weights to differ and only regularizes them to keep them related. However, it has been shown in~\cite{Rad2018} that the adapted features are not accurate enough for 3D pose estimation.
 
Differently, \cite{Zakharov2018} transfers real depth images to clean synthetic-looking depth images.  However, this requires extensive hand-crafted depth image augmentation to create artificial real-looking depth images during training, and modeling the noise of real depth cameras is difficult in practice.  \cite{Tobin2017} proposed to randomize the appearance of objects and rendering parameters during training, in order to improve generalization of the trained model to real-world scenarios.  However, this requires significant engineering effort and there is no guarantee that these randomized renderings cover all the visual appearances of a real-world scene.

Several works~\cite{Hoffman2016,Song2017} propose a fusion of features from different domains.  \cite{Hoffman2016} fuses color and depth features by using labeled depth images for a few categories and adapts a color object detector for a new category such that it can use depth images in addition to color images at test time. \cite{Song2017} propose a combination method that selects discriminative combinations of layers from the different source models and target modalities and sums the features before feeding them to a classifier.  However, both works require annotated images in all domains.  \cite{Huang2017} uses a shared network that utilizes one modality in both source and target domain as a bridge between the two modalities, and an additional network that preserves the cross-modal semantic correlation in the target domain.  However, they require annotations in both domains, whereas we only require annotations in one, \ie the synthetic, domain that are much easier to acquire.

When comparing our work to these related works on domain transfer learning, these methods either require annotated examples in the target domain~\cite{Mueller2018,Huang2017,Hoffman2016,Song2017,Rad2018,Cai2018}, are restricted to two domains~\cite{Gupta2016,Song2017,Hoffman2016,Rad2018,Cai2018}, or require significant engineering~\cite{Zakharov2018,Tobin2017}. By contrast, our method does not require any annotations in the target domain, \ie color images, and can be only trained on synthetically rendered depth images that are easy to generate, and the domain transfer is trained from real data that can be easily acquired using a commodity RGB-D camera.


\section{Method}

\newcommand{\bff}{{\bf f}}
\newcommand{\bx}{{\bf x}}
\newcommand{\by}{{\bf y}}
\newcommand{\calT}{\mathcal{T}}
\newcommand{\calD}{\mathcal{D}}
\newcommand{\calC}{\mathcal{C}}
\newcommand{\calR}{\mathcal{R}}
\newcommand{\calS}{\mathcal{S}}
\newcommand{\calL}{\mathcal{L}}
\newcommand{\RGBD}{\texttt{RGB-D}}
\newcommand{\FM}{\texttt{FM}}

Given a 3D model of a target object, it is easy to generate a training set made of many depth images of the object under different 3D poses. Alternatively, we can use an existing dataset of labeled depth images. We use this training set to train a first network to extract features from such depth images, and a second network, the \textit{pose estimator}, to predict the 3D pose of an object, given the features extracted from a depth image. Because it is trained on many images, the pose estimator performs well, but only on depth images.

\begin{figure}[t!]
  \centering
  \includegraphics[width=0.92\linewidth]{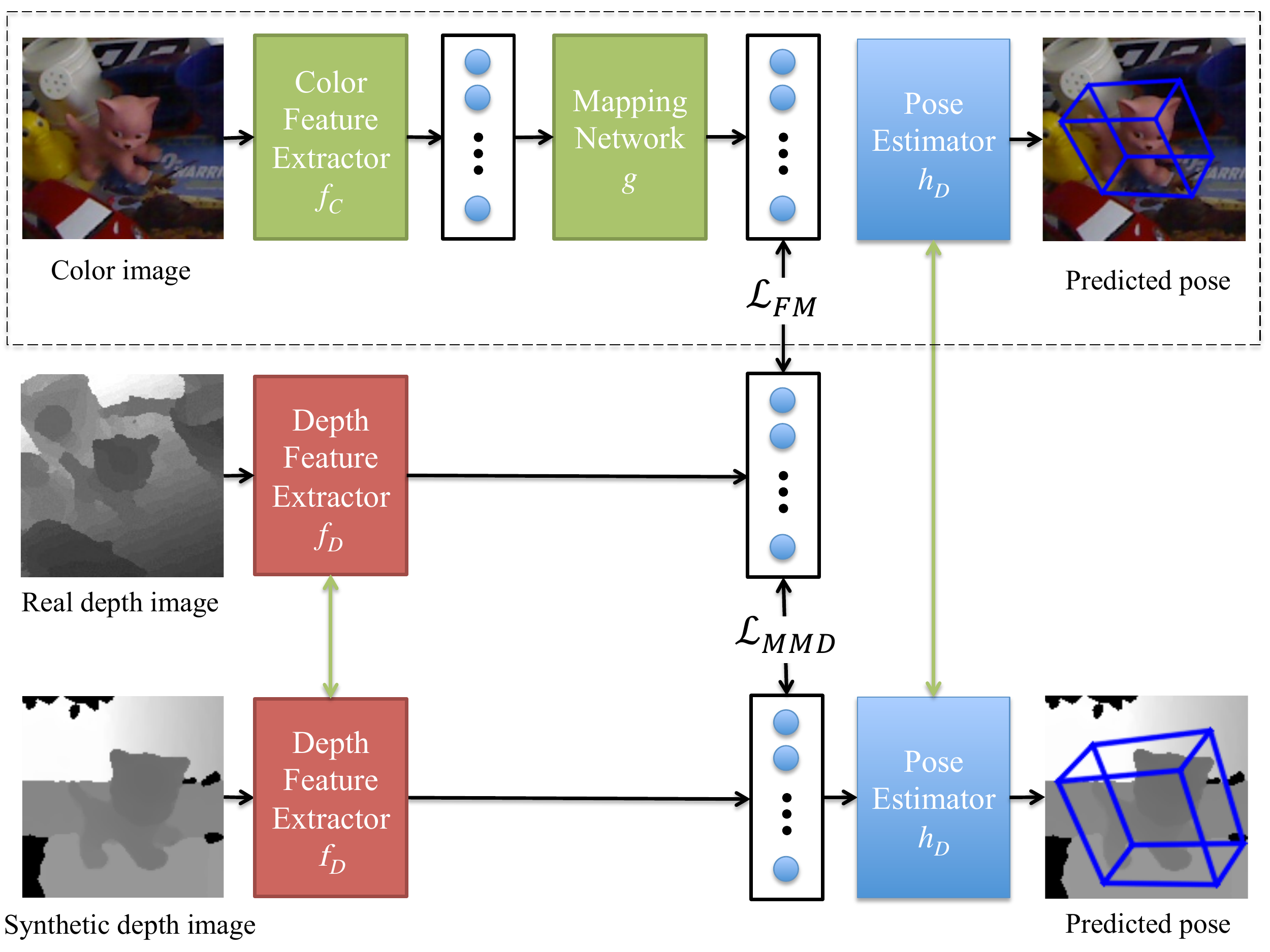}
  \caption{Detailed overview of our approach. It consists of three
    data streams, one for each domain. The lower two streams take
    depth images as input, \ie synthetic and real depth images, respectively, and extract features using the network $f_\calD$. The upper stream takes color images as input and uses the mapping network $g$ to map the color features to the depth features used for pose prediction with network $h_\calD$. The parameters of the depth feature extractor $f_\calD$ and pose predictor $h_\calD$ are shared between the synthetic and real depth image (green arrows). Between the synthetic and the real depth feature we have the MMD loss $\calL_{MMD}$ and between the real color and real depth features we use the feature mapping loss $\calL_{FM}$. For inference at test time, we use only the upper stream within the dashed lines that takes a real color image as input, extracts features using the network $f_\calC$, maps these features to the depth space using $g$, and uses the pose estimator $h_\calD$ to predict the 3D pose.}
  \label{fig:method}
\end{figure}

To apply the pose estimator network to color images, we train a network to map features extracted from color images to features extracted from depth images, as was done in \cite{Rad2018} between real and synthetic images.  To learn this mapping, we capture a set of pairs of color and depth images that correspond to the same view, using an RGB-D camera.  In order to handle the domain gap between the real and synthetic depth images of two training sets, we apply the Maximum Mean Discrepancy~(MMD)~\cite{Gretton08}, which aims to map features of each training set to a Reproducing Kernel Hilbert Space~(RKHS) and then minimizes the distance between the means of the mapped features of the two training sets.  An overview of the proposed method is shown in Fig.~\ref{fig:method}.

\subsection{Learning the Mapping}

More formally, let $\calT^\calS=\{(\bx_i^\calS,\by_i)\}_i$ be a training set of synthetically rendered depth images $\bx_i^\calS$ using a 3D renderer engine under 3D poses $\by_i$.  A second training set $\calT^\RGBD=\{(\bx_i^\calR,\bx_i^\calD)\}_i$ consists of pairs of color images $\bx_i^\calR$, and their corresponding depth images $\bx_i^\calD$.  We jointly train four networks: the feature extractor for depth images $f_\calD$, the feature extractor for color images $f_\calC$, the pose estimator $h_\calD$, and the feature mapping network $g$, on the training sets $\calT^\calD$ and $\calT^\RGBD$.

We optimize the following loss function over the parameters of networks 
$f_\calD$, $h_\calD$, $f_\calC$, and $g$ as:
\begin{equation}
\label{eq:loss}
  \begin{array}{ll}
&  \calL(\theta_D, \theta_h, \theta_C, \theta_g;\calT^\calS, \calT^\RGBD) = \\
    &  \calL_P(\theta_D, \theta_h; \calT^\calS) + \beta \calL_{FM}(\theta_D, \theta_C, \theta_g; \calT^\RGBD)  + \gamma \calL_{MMD}(\theta_D;\calT^\calS,\calT^\RGBD) \> ,
    \end{array}
\end{equation}
where $\theta_D$, $\theta_h$, $\theta_C$, and $\theta_g$ are the parameters of 
networks $f_\calD$, $h_\calD$, $f_\calC$, and $g$, respectively. The losses $\calL_P$ for the pose, $\calL_{FM}$ for the feature mapping between color and depth features, and $\calL_{MMD}$ for the MMD between synthetic and real depth images are 
weighted by  meta parameters $\beta$ and $\gamma$.\\
$\calL_P$ is the sum of the errors for poses predicted from depth images:
\begin{equation}
  \calL_P(\theta_D, \theta_h; \calT^\calS) = \sum_{(\bx_i^\calS, \by_i) \in \calT^\calS} \lVert h_\calD(f_\calD(\bx_i^\calS;
  \theta_D);\theta_h)-\by_i \rVert^2 \> .
\end{equation}\\
$\calL_{FM}$ is the loss used to learn to map features extracted from depth images to features extracted from their corresponding color images:
\begin{equation}
  \calL_{FM}(\theta_D, \theta_C, \theta_g; \calT^\RGBD) = \sum_{(\bx_i^\calR, \bx_i^\calD) \in \calT^\RGBD} \lVert g(f_\calC(\bx_i^\calR; \theta_C); \theta_g) - f_\calD(\bx_i^\calD; \theta_D) \rVert^2 \> .
\end{equation}\\
Finally, $\calL_{MMD}$ is the Maximum Mean Discrepancy~\cite{Gretton08} loss to minimize the domain shift between the
distribution of features extracted from real and synthetic depth images of these training sets:
\begin{equation}
  \begin{array}{lll}
    \calL_{MMD}(\theta_D;\calT^\calS,\calT^\RGBD) = \Big\lVert & \frac{1}{|\calT^\RGBD|} \sum_{\bx_i^\calD \in \calT^\RGBD} \phi(f_\calD(\bx_i^\calD;\theta_D)) \;- &\\
    &\frac{1}{|\calT^\calS|} \sum_{\bx_i^\calS \in \calT^\calS} \phi(f_\calD(\bx_i^\calS; \theta_D)) & \Big\rVert^2 \> ,
  \end{array}
\end{equation}
where $\phi(\cdot)$ denotes the mapping to kernel space, but the exact mapping is typically unknown in practice. By applying the kernel trick, this rewrites to:
\begin{equation}
\arraycolsep=1.4pt\def\arraystretch{1.4}
  \begin{array}{ll}
\calL_{MMD}(\theta_D;\calT^\calS,\calT^\RGBD) = & \frac{1}{{|\calT^\RGBD|}^2} \sum_{i,i'} k(f_\calD(\bx_i^\calD;\theta_D), f_\calD(\bx_{i'}^\calD;\theta_D))  \\
 & - \frac{2}{|\calT^\RGBD| |\calT^\calS|}\sum_{i,j} k(f_\calD(\bx_i^\calD;\theta_D), f_\calD(\bx_j^\calS;\theta_D))  \\
 & + \frac{1}{{|\calT^\calS|}^2}\sum_{j,j'} k(f_\calD(\bx_j^\calS;\theta_D), f_\calD(\bx_{j'}^\calS;\theta_D)) \,\, ,
  \end{array}
\end{equation}
where $k(\cdot, \cdot)$ denotes a kernel function. In this work, we implement $k(\cdot, \cdot)$ as an RBF kernel, such that 
\begin{equation}
k(\mathbf{x},\mathbf{y})=\mathrm{e}^{-\frac{\|\mathbf{x} - \mathbf{y}\|^{2}}{2\sigma^{2}}} \,\, ,
\end{equation}
where we select the bandwidth $\sigma = 1$. Note that our method is not sensitive to the exact value of $\sigma$.

At run-time,  given a real color image $\bx^\calR$,  we extract its features in color space and map them to the 
depth feature space by the networks $f_\calC$ and $g$, respectively, and then use the pose estimator
$h_\calD$ to predict the 3D pose $\hat{\by}$
of the object:
\begin{equation}
    \hat{\by} = h_\calD(g(f_\calC(\bx^\calR))) \,\, .
\end{equation}

\subsection{Network Details and Optimization}

\subsubsection{3D Object Pose Estimation}
For the depth feature extraction network $f_\calD$, we use a network architecture similar to the 50-layer
Residual Network~\cite{He16}, and remove     the   Global   Average
Pooling~\cite{He16} as done in~\cite{Rad2018,Oberweger17,Mueller17}. The convolutional layers are followed by
two fully-connected layers of 1024 neurons each. The pose estimator $h_\calD$ consists of 
one single fully-connected layer with
16 outputs, which correspond to the 2D projections of the 8 corners of objects' 3D bounding box~\cite{Rad17}.
The 3D pose can then be computed from these 2D-3D correspondences with a P$n$P algorithm.

In order to do a fair comparison to~\cite{Rad2018,Rad17} we use the same feature extractor, which consists of the first 10 pretrained convolutional layers of the VGG-16 network~\cite{Simonyan14}
and  two fully-connected layers with 1024 neurons for the color feature extractor $f_\calC$.

\subsubsection{3D Hand Pose Estimation}
The architectures of $f_\calD$ and $h_\calD$ are the same as the ones used for 3D object pose estimation, 
except $h_\calD$ outputs 3 values for  each of the 21  joints, 63 in total. Additionally, we add a 3D pose prior~\cite{Oberweger17} as a bottleneck layer to the pose estimator network $h_\calD$, which was shown to efficiently constrain the 3D hand poses and also gives better performance in our case.

For the color feature extractor $f_\calC$, we use the same architecture as the depth feature extractor, which makes the feature extractor comparable to the one used in~\cite{Mueller2018,Spurr2018}.

\subsubsection{Mapping Network and Optimization}
Following \cite{Rad2018}, we use a two Residual blocks~\cite{He15} network $g$ for mapping the features of size 1024 from color space to depth space. Each fully-connected layer within the Residual block has 1024 neurons.

In practice, we use $\beta = 0.02$ and $\gamma = 0.01$ for the meta parameters of the objective function in Eq.~\ref{eq:loss} for all our experiments.  We first pretrain $f_\calD$ and $h_\calD$ on the synthetic depth dataset $\calT^\calS$.  We also pretrain $f_\calC$ by predicting depth from color images~\cite{Eigen14}.  Pretraining is important in our experiments for improving convergence.  We then jointly train all the networks together using the ADAM optimizer~\cite{Kingma15} with a batch size of 128 and an initial learning rate of $10^{-4}$.


\section{Evaluation}
  
In this section, we evaluate our method on two different 3D pose
estimation problems.  We apply our method first on 3D rigid object
pose estimation, and then on 3D hand pose estimation, both from color images only.

\subsection{3D Object Pose Estimation from Color Images}

We use the LINEMOD dataset~\cite{Hinterstoisser12b} for benchmarking 3D object pose estimation. It consists  of 13
 texture-less objects, each registered with about 1200 real color images and corresponding depth images under different viewpoints, 
 and provided with the corresponding ground truth poses.
For evaluating 3D object pose estimation methods using only color images, \cite{Xiang2018,Rad2018,Rad17,Brachmann16} use 15\% of the images of the LINEMOD dataset for training and the rest for testing.  This amounts to about 180 images per object for training, which had to be registered in 3D with the help of markers.  In contrast to these methods, our approach does {\it not} require any labeled color image. Instead, it uses pairs of real images and depth images to learn mapping the features from color space to depth space.

\begin{table}
\begin{center}
\resizebox{\textwidth}{!}{
\begin{tabular}{l|ccc|ccc|ccc}
\toprule
Detection & \multicolumn{3}{c}{Ground Truth Detection} & \multicolumn{6}{|c}{Real Detection}\\
\midrule
Metric & \multicolumn{3}{c}{2D Projection~\cite{Brachmann16}} & \multicolumn{3}{|c}{2D Projection~\cite{Brachmann16}}  & \multicolumn{3}{|c}{ADD~\cite{Hinterstoisser12b}} \\
Method & BB8~\cite{Rad17}  & Feature Mapping~\cite{Rad2018} & Ours & SSD-6D~\cite{Kehl17} & ~~~\cite{sundermeyer2018implicit}~~~~ & Ours & SSD-6D~\cite{Kehl17} & ~~~\cite{sundermeyer2018implicit}~~~~ & Ours \\
\midrule
Ape           &  94.0  & 96.6       & {\bf 97.3} &       3.5 & 36.4 & {\bf 96.9} &       2.6 &  4.0 & {\bf 19.8}\\     
Bench Vise    &  90.0  & {\bf 96.3} & 92.7 &             0.9 & 30.5 & {\bf 88.6} &      15.1 & 20.9 & {\bf 69.0}\\
Camera        &  81.7  & {\bf 94.8} & 83.4 &             1.0 & 56.0 & {\bf 77.4} &       6.1 & 30.5 & {\bf 37.6}\\
Can           &  94.2  & {\bf 96.6} & 93.2 &             3.0 & 49.1 & {\bf 91.3} &      27.3 & 35.9 & {\bf 42.3}\\
Cat           &  94.7  & 98.0       & {\bf 98.7} &       9.1 & 59.3 & {\bf 98.0} &       9.3 & 17.9 & {\bf 35.4}\\
Driller       &  64.7  & {\bf 83.3} & 75.7 &             1.4 & 16.7 & {\bf 72.2} &      12.0 & 24.0 & {\bf 54.7}\\
Duck          &  94.4  & {\bf 96.3} & 95.5 &             1.2 & 51.0 & {\bf 91.8} &       1.3 &  4.9 & {\bf 29.4}\\
Egg Box       &  93.5  & 96.1       & {\bf 97.1} &       1.5 & 73.5 & {\bf 92.0} &       2.8 & 81.0 & {\bf 85.2}\\
Glue          &  94.8  & 96.9       & {\bf 97.3} &      11.0 & 78.3 & {\bf 92.4} &       3.4 & 45.5 & {\bf 77.8}\\
Hole Puncher~ &  87.2  & 95.7       & {\bf 97.2} &       2.8 & 48.2 & {\bf 96.8} &       3.1 & 17.6 & {\bf 36.0}\\
Iron          &  81.0  & {\bf 92.3} & 88.8 &             1.9 & 32.1 & {\bf 85.9} &      14.6 & 32.0 & {\bf 63.1}\\
Lamp          &  76.2  & 83.5       & {\bf 84.8} &       0.5 & 30.8 & {\bf 81.8} &      11.4 & 60.5 & {\bf 75.1}\\
Phone         &  70.6  & 88.2       & {\bf 90.0} &       5.3 & 53.3 & {\bf 85.2} &       9.7 & 33.8 & {\bf 44.8}\\
\midrule
Average      &  85.9   & {\bf 93.4} & 91.7       &       3.3 & 47.3 & {\bf 88.5} &       9.1 & 28.7 & {\bf 51.6}\\
\bottomrule
\end{tabular}}
\vspace{2mm}
\caption{Evaluation on the LINEMOD dataset~\cite{Hinterstoisser12b}.
  The left part evaluates the impact of our proposed approach, where all methods predict the 3D object pose using the 2D
  projections   of  the objects'   3D   bounding   box~\cite{Rad17}, given 
  the ground truth 2D object center without using pose refinement. 
  Both BB8~\cite{Rad17} and Feature Mapping~\cite{Rad2018} use 
  annotated color images, while our method achieves better performance than BB8 and similar performance 
  to Feature Mapping
  without using any annotated color images at all.
  The middle and right parts show comparison of different pose estimation methods without using pose refinement, where no annotated color image is used for training. Our approach performs best.  
   }
   \label{tbl:result_1}
\end{center}
\end{table}

In order to do a fair comparison and not learn any context of the scene, 
we extract the objects from both color images and depth images for generating the
training set $\calT^\RGBD$ .  We follow the protocol of \cite{Rad17} to augment
the training data by rescaling the target object, adding a small pixel shift from the
center of the image window and superimpose it on a random
background. We pick random backgrounds from the RGB-D
dataset~\cite{Lai2011} as they provide color images together with corresponding depth images.

For generating the training set $\calT^\calS$, given the CAD model of the target object,
we randomly sample poses from the  upper hemisphere of the object,
within a range of $[-45^\circ, +45^\circ]$ for the in-plane rotation
and a camera distance  within a range of $[65cm, 115cm]$. We also superimpose the 
rendered objects on a random depth background picked from the RGB-D
dataset~\cite{Lai2011}. We apply a $5 \times 5$ median filter to mitigate the 
noise along the object boundaries. For both training sets $\calT^\RGBD$ and $\calT^\calS$,
we use image windows of size $128\times 128$, and normalize them to the range of $[-1, +1]$.

To evaluate the impact of our approach, we first compare it to \cite{Rad17} and 
\cite{Rad2018}, which, similar to us, predict the 2D projection of the 3D object
bounding box, followed by a P$n$P algorithm to estimate the 3D pose. The left part of Table~\ref{tbl:result_1}
shows a comparison with these methods on the LINEMOD dataset by using 15\% of real images for training and the ground truth 2D object center. We use the widely used
2D Projection metric~\cite{Brachmann16} for comparison. We significantly
outperform~\cite{Rad17} and achieve similar performance to~\cite{Rad2018}, which is the current state-of-the-art on the LINEMOD dataset. Most notably, we do not require any annotations for the color images.
By using all the available pairs of real images and depth images, our approach performs
with an accuracy of 95.7\%. This shows that our
approach almost eliminates the needs of the expensive task of annotating, simply by capturing
data using an RGB-D camera.  

We further compare to the approach of~\cite{Kehl17}\footnote{We used their public code to obtain accuracies on 2D Projection and ADD metrics.} without the extensive refinement step, which uses only synthetic color images for training. They obtain an accuracy of 3.3\% on the 2D Projection metric. \cite{Rad17} trained only on synthetic color images also performs poorly with an accuracy of 12\% on the same metric.  This shows that while synthetic color images do not resemble real color images for training 3D pose estimation methods, our approach can effectively transfer features between color images and synthetic depth images. Although the domain randomization of \cite{sundermeyer2018implicit} helps to increase the accuracy using synthetically  generated images and generalize to different cameras, it is still not enough to bridge the domain gap. The comparisons are shown in Table~\ref{tbl:result_1}.

Finally, we  evaluate the domain adaption technique of~\cite{Ganin15} that aims to learn  invariant features with respect
to the shift between the color and depth domains. However, this performs with an accuracy of 2\% using the 2D Projection metric, which shows that
although this technique helps for general applications such as classification, the features are not well suited for 3D pose estimation.

\subsection{3D Hand Pose Estimation from Color Images}

We use the Stereo Hand Pose Benchmark~(STB)~\cite{Zhang2016} and the Rendered Hand Dataset~(RHD)~\cite{Zimmermann2017} for training and testing our approach for 3D hand pose estimation.  The STB dataset contains 6 sequences each containing 1500 images of a stereo camera, and an RGB-D camera. It shows an user performing diverse hand gestures in front of different backgrounds. Each image is annotated with the corresponding 3D hand pose with 21 joints.  The RHD dataset contains over 40k synthetically rendered images of humans performing various hand articulations.  It consists of pairs of depth and color images, each with hand segmentation and 3D pose annotations of the same 21 joints.

\begin{figure}[t!]
	\centering
	\includegraphics[width=0.85\linewidth]{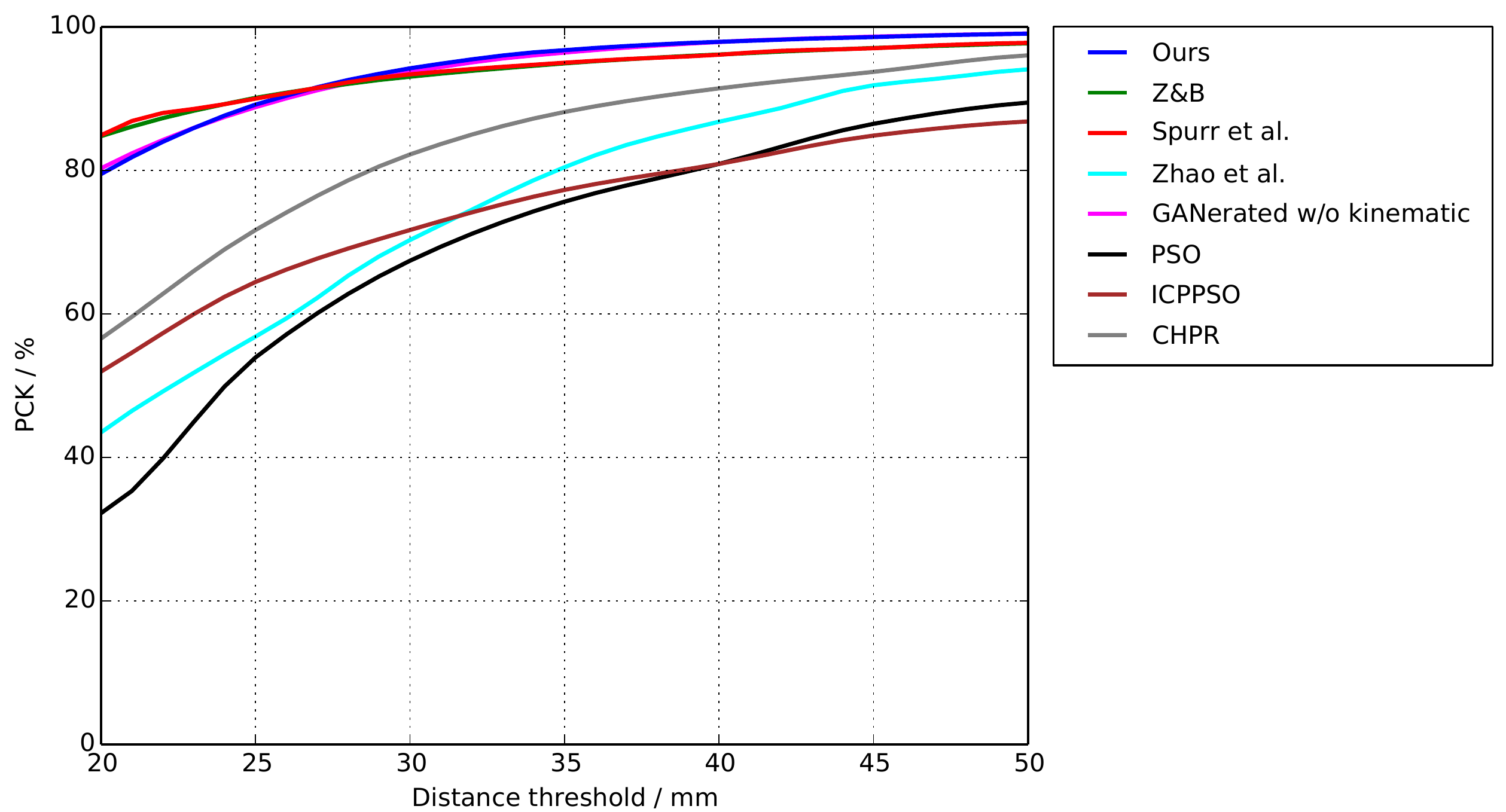}
	\caption{3D PCK curves for comparison to state-of-the-art 3D hand pose estimation methods on the STB dataset~\cite{Zhang2016}. Note that all other approaches use annotated color images for training, whereas we do not use any annotations for the color images.}
	\label{fig:stb_pck}
\end{figure}

We follow the protocol of~\cite{Zimmermann2017,Mueller2018,Spurr2018}, which use the first two sequences from the STB dataset for testing. The remaining ten sequences from the STB dataset together with the RHD dataset are used for the training set $\calT^\RGBD$, since they both contain aligned depth and color images. Creating synthetic depth maps for
hands is a relatively simple problem. For generating the training set $\calT^\calS$ we use the publicly available 3D hand model of~\cite{Tompson14b} to render synthetic depth images of a hand. We use 5M synthetic images of the hand that are rendered online during training from poses of the NYU 3D hand pose dataset~\cite{Tompson14b} perturbed with randomly added articulations. Furthermore, \cite{Zimmermann2017,Mueller2018} align their 3D prediction to the ground truth wrist which we also do for comparison.

We use the pipeline provided by~\cite{Oberweger17} to preprocess the depth images: It crops a $128\times 128$ patch around the hand location, and normalizes its depth values to the range of $[-1, +1]$. For the color image we also crop a $128\times 128$ patch around the corresponding hand location and subtract the mean RGB values. When a hand segmentation mask is available, such as for the RHD dataset~\cite{Zimmermann2017}, we superimpose the segmented hand on random color backgrounds from the RGB-D dataset~\cite{Lai2011}. During training, we randomly augment the hand scale, in-plane rotation, and 3D translation, as done in~\cite{Oberweger17}.

We compare to the following methods: GANerated~\cite{Mueller2018}\footnote{The results reported in the paper~\cite{Mueller2018} are tracking-based and include an additional inverse kinematics step. In order to make their results more comparable to ours, we denote results predicted for each frame separately without inverse kinematics kindly provided by the authors.}, which uses a GAN to adapt synthetic color images for training a CNN; Z\&B~\cite{Zimmermann2017}, which uses a learned prior to lift 2D keypoints to 3D locations and the similar approach of Zhao~\etal~\cite{Zhao2016}; Zhang~\etal~\cite{Zhang2016}, which use stereo images to estimate depth and apply a depth-based pose estimator with variants denoted PSO, ICPPSO, and CHPR; Spurr~\etal~\cite{Spurr2018}, which project color images to a common embedding that is shared between images and 3D poses.

Fig.~\ref{fig:stb_pck} shows the Percentage of Correct Keypoints~(PCK) over different error thresholds, which is the most common metric on the STB dataset~\cite{Zhang2016,Zimmermann2017,Mueller2018,Spurr2018}. This metric denotes the average percentage of predicted joints below an Euclidean distance from the ground truth 3D joint location. While all methods that we compare to require annotations for color images, we can achieve comparable results without annotations of color images.

\begin{figure}
	\centering
	\begin{tabular}{cccc}
	Color image & \begin{tabular}{@{}c@{}}Pose prediction on \\ real depth image\end{tabular} & \begin{tabular}{@{}c@{}}Pose prediction on \\ predicted depth image\end{tabular} & \begin{tabular}{@{}c@{}}Our approach on \\ color image\end{tabular} \\
    \includegraphics[width=0.17\linewidth]{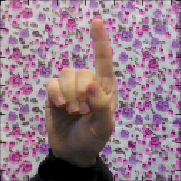} &
    \includegraphics[width=0.17\linewidth]{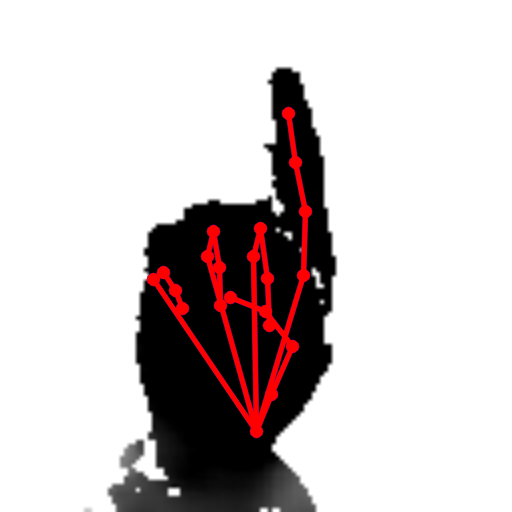} &
    \includegraphics[width=0.17\linewidth]{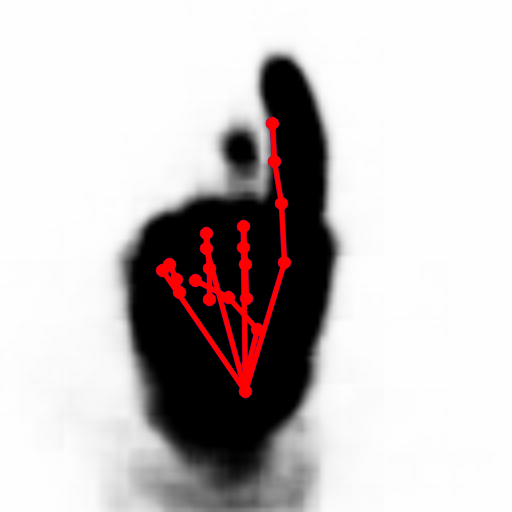} &
    \includegraphics[width=0.17\linewidth]{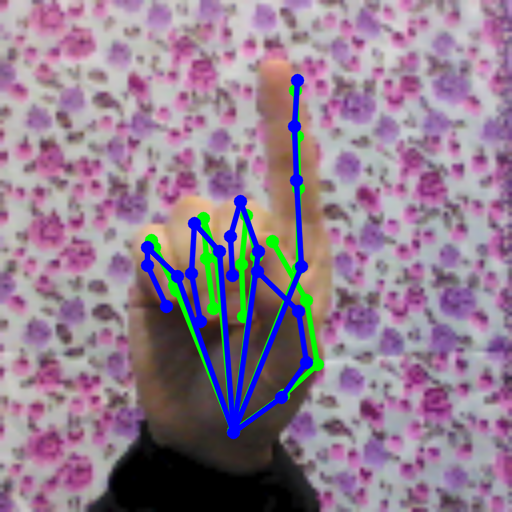} \\
    \midrule
    \includegraphics[width=0.17\linewidth]{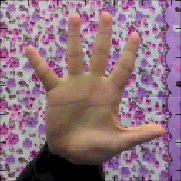} &
    \includegraphics[width=0.17\linewidth]{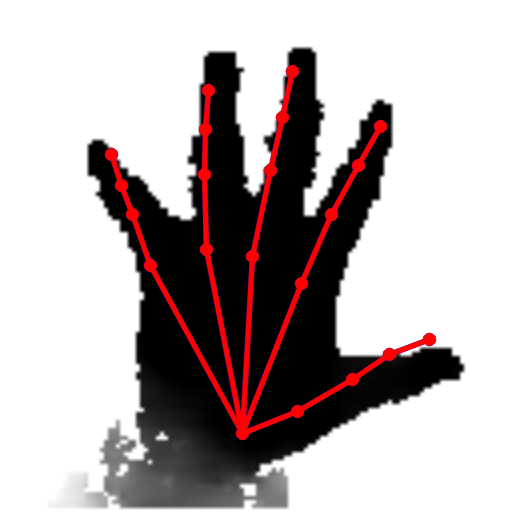} &
    \includegraphics[width=0.17\linewidth]{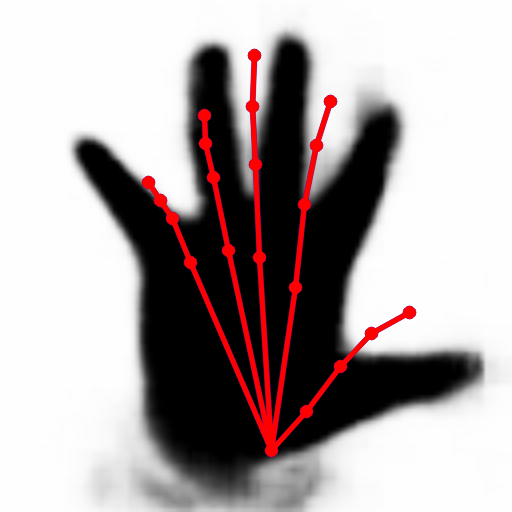} &
    \includegraphics[width=0.17\linewidth]{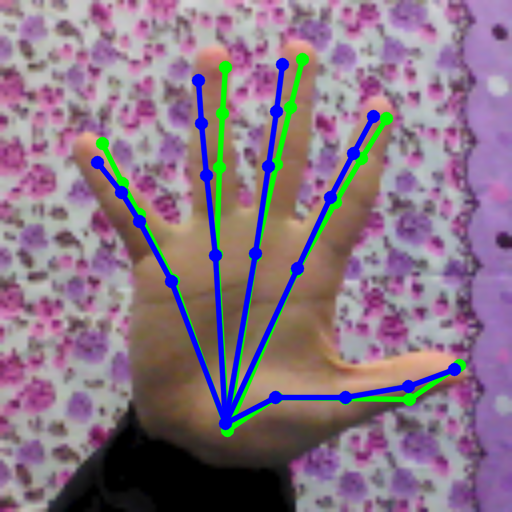} \\
    (a) & (b) & (c) & (d) \\
    \end{tabular}
	\caption{We use the paired color and depth images shown in (a) and (b) to predict depth from color images~\cite{Eigen14} shown in (c). We further apply a 3D pose estimator~\cite{Oberweger17} on these images. These predictions from the predicted and real depth
          images are shown in (b) and (c), respectively. Although the predicted depth images look visually close, the accuracy of the estimated 3D pose is significantly worse compared to using the real depth images. The results from our method are shown in (d) and provide a significantly higher accuracy. Our predictions are shown in blue and the ground truth in green.}
	\label{fig:depth_based_pose}
\end{figure}

3D hand pose estimation methods work very well on depth images~\cite{Oberweger17,Mueller17}. Since we have paired color and depth images, we can train a CNN to predict the depth image from the corresponding color image~\cite{Eigen14}. Since the pose estimator works on cropped image patches, we only use these cropped image patches for depth prediction, which makes the task easier. We then use the predicted depth image for a depth-based 3D hand pose estimator with the pretrained model provided by the authors~\cite{Oberweger17}. Although this approach also does not require any annotations of color images, our experiments show that this performs significantly worse on the STB dataset compared to ours. The 3D pose estimator gives an average Euclidean joint error of 17.6mm on the real depth images and 39.8mm on the predicted depth images. We show a qualitative comparison in Fig.~\ref{fig:depth_based_pose}.

\subsection{Qualitative Results}
\begin{figure}
	\centering
	\begin{tabular}{cccc}
    \includegraphics[width=0.22\linewidth]{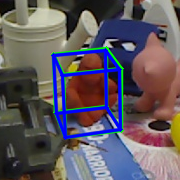} &
    \includegraphics[width=0.22\linewidth]{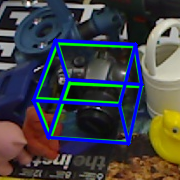} &
    \includegraphics[width=0.22\linewidth]{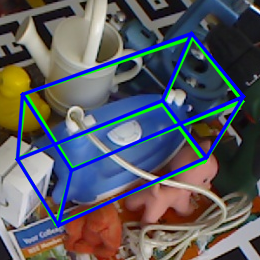} &
    \includegraphics[width=0.22\linewidth]{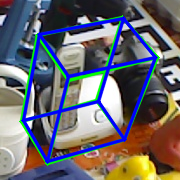} \\

    \midrule
    \includegraphics[width=0.22\linewidth]{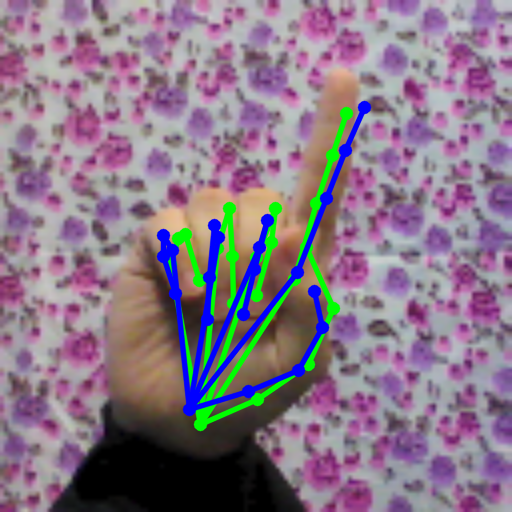} &
    \includegraphics[width=0.22\linewidth]{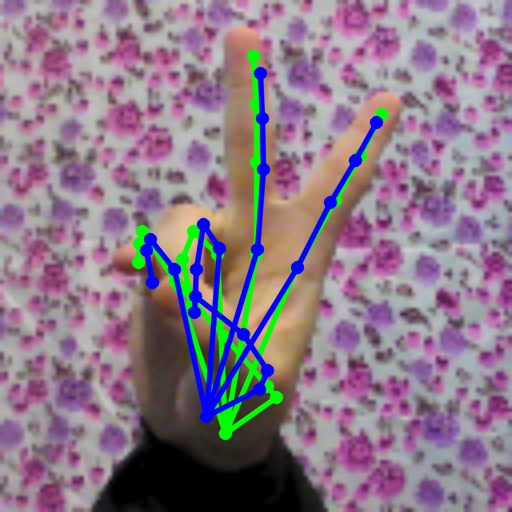} &
    \includegraphics[width=0.22\linewidth]{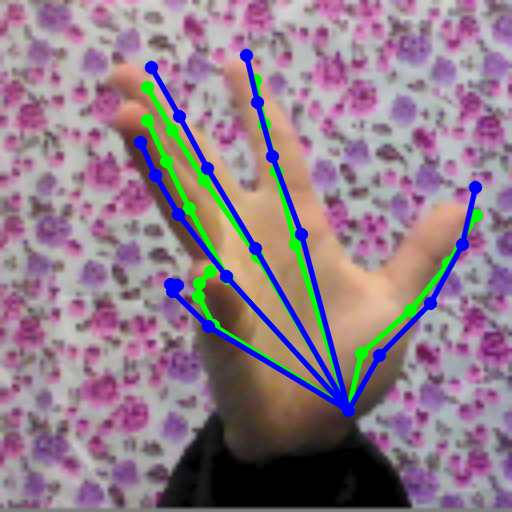} &
    \includegraphics[width=0.22\linewidth]{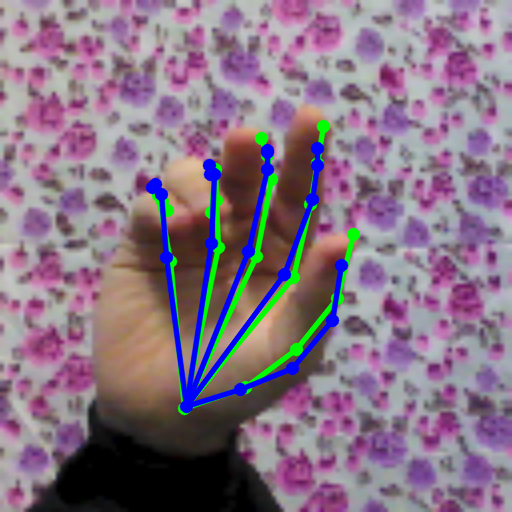}\\

    \midrule
    \includegraphics[width=0.22\linewidth,clip]{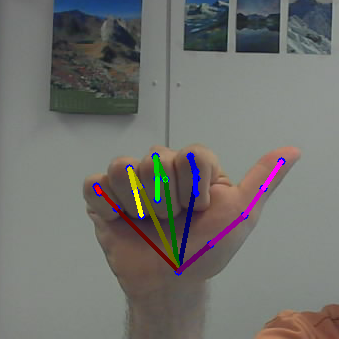} &
    \includegraphics[width=0.22\linewidth,clip]{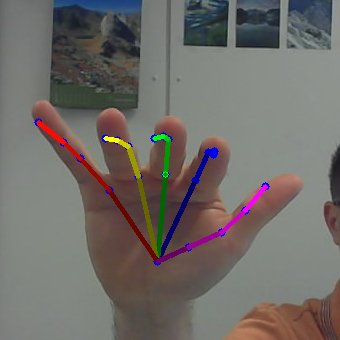} &
    \includegraphics[width=0.22\linewidth,clip]{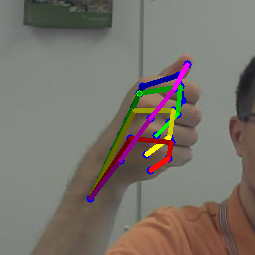} &
    \includegraphics[width=0.22\linewidth,clip]{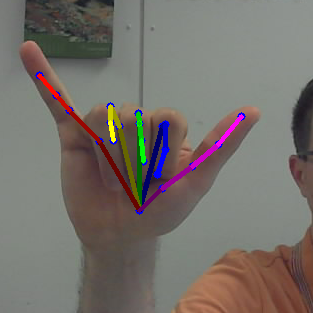} \\
    \includegraphics[width=0.22\linewidth,clip]{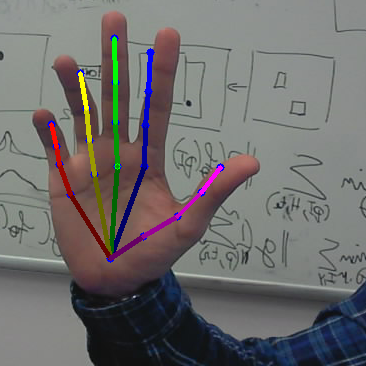} &
    \includegraphics[width=0.22\linewidth,clip]{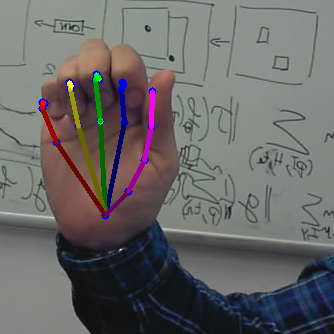} &
    \includegraphics[width=0.22\linewidth,clip]{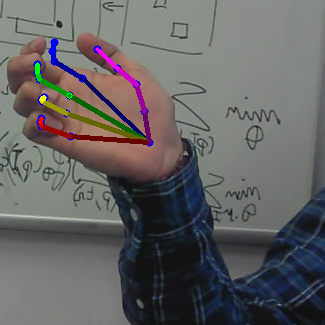} &
    \includegraphics[width=0.22\linewidth,clip]{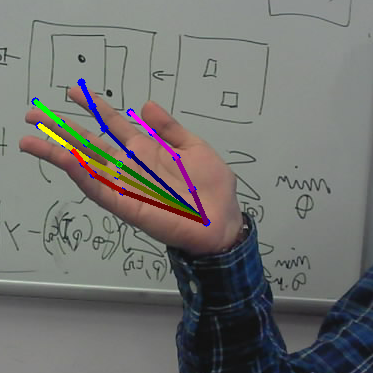} \\
    \end{tabular}
	\caption{Qualitative results of our method for 3D rigid object pose estimation on the LINEMOD dataset~\cite{Hinterstoisser12b} (top row), and 3D hand pose estimation on the STB dataset~\cite{Zhang2016} (middle row). Green denotes ground truth and blue corresponds to the predicted pose. We applied our trained network on real world RGB images of different users to estimate the 3D hand joint locations (bottom rows).}
	\label{fig:qualitative}
\end{figure}

We show some qualitative results of our method for 3D object pose estimation and 3D hand pose estimation in Fig.~\ref{fig:qualitative}. These examples show our approach predicts very close pose to the ground truth.

\begin{figure}[t!]
	\centering
	\begin{tabular}{cccc}
    \includegraphics[width=0.2\linewidth]{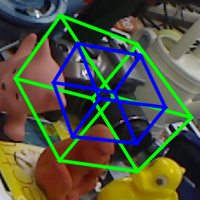} &
    \includegraphics[width=0.2\linewidth]{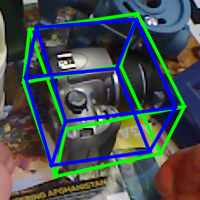} &
    \includegraphics[width=0.2\linewidth]{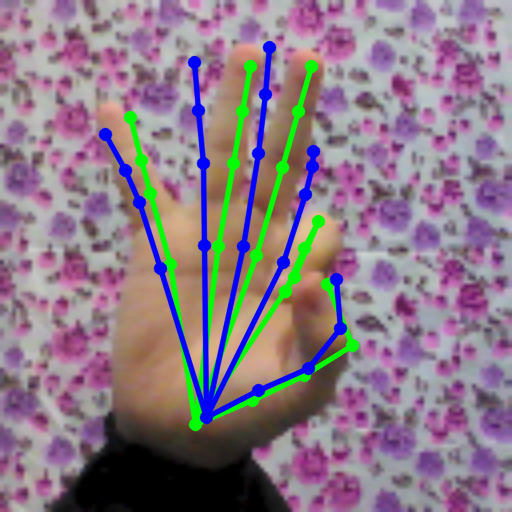} &
    \includegraphics[width=0.2\linewidth]{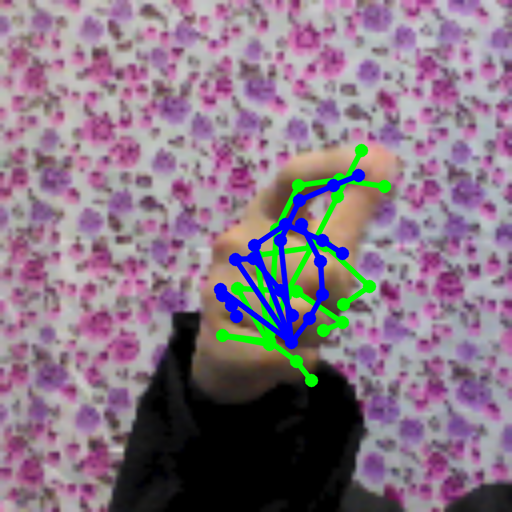} \\
    (a) & (b) & (c) & (d)\\
    \end{tabular}

	\caption{Some failure cases for 3D object pose estimation due to (a) partial occlusion, (b) not generalizing to every poses because of 
	 lack of corresponding color and depth images in the training set. Failure cases for 3D hand pose estimation due to (c) misalignment/confusion of the fingers, (d) severe self occlusion.}
	\label{fig:failure}
\end{figure}

Fig.~\ref{fig:failure} illustrates some failure cases that occur due to the challenges of the test sets, such as partial occlusion that can easily be handled by training the networks with partially occluded examples, or missing poses in the paired dataset $\calT^\RGBD$ that can be simply resolved by capturing additional data with an RGB-D camera.

\subsection{Computation Times}
All experiments are implemented using Tensorflow and run on an Intel Core i7 3.3GHz desktop with a Geforce TITAN X.  Given an image window extracted around the object, our approach takes 3.2ms for 3D object pose estimation to extract color features, map them to the depth feature space, and estimate the 3D pose.  For 3D hand pose estimation, it takes 8.6ms. Training takes about 10 hours in our experiments.


\section{Conclusion}

In this work we presented a novel  approach for 3D pose estimation from color images,
without requiring labeled color images.
We showed that a pose estimator can be trained on a large number of synthetic depth images, and 
at run-time, given a color image, we can map its features from color space to depth space. 
We showed that this mapping
between the two domains can easily be learned by having corresponding color and depth images 
captured by a commodity RGB-D camera.
Our approach is simple, general, and can be applied to different application domains,
such as 3D rigid object pose estimation and 3D hand pose estimation. While for these tasks our approach achieves performances 
comparable to state-of-the-art methods, it does not require any annotations for the color images.

\paragraph{Acknowledgement:} This work was supported by the Christian Doppler Laboratory for Semantic 3D Computer Vision, funded in part by Qualcomm Inc. We would like to thank Franziska M\"uller and Martin Sundermeyer for kindly providing additional evaluation results.
Prof. V. Lepetit is a senior member of the \emph{Institut Universitaire de France}~(IUF).
\clearpage
{
\setstretch{0.98}
\bibliographystyle{splncs}
\bibliography{short,biblio,vision}
}

\end{document}